\documentclass{article}
\usepackage{spconf,amsmath,graphicx,multirow, xcolor, amssymb,subcaption,caption,microtype}

\title{MIXED-ATTENTION AUTO ENCODER FOR MULTI-CLASS INDUSTRIAL ANOMALY DETECTION}
\name{Jiangqi Liu, Feng Wang$^{\star}$}
\vspace{-.15in}
\address{Shanghai Key Laboratory of Multidimensional Information Processing\\ School of Computer Science and Technology \\East China Normal University, Shanghai, China}

\begin{document}
%
\maketitle
\begin{abstract}
Most existing methods for unsupervised industrial anomaly detection train a separate model for each object category. This kind of approach can easily capture the category-specific feature distributions, but results in high storage cost and low training efficiency. In this paper, we propose a unified mixed-attention auto encoder (MAAE) to implement multi-class anomaly detection with a single model. To alleviate the performance degradation due to the diverse distribution patterns of different categories, we employ spatial attentions and channel attentions to effectively capture the global category information and model the feature distributions of multiple classes. Furthermore, to simulate the realistic noises on features and preserve the surface semantics of objects from different categories which are essential for detecting the subtle anomalies, we propose an adaptive noise generator and a multi-scale fusion module for the pre-trained features. MAAE delivers remarkable performances on the benchmark dataset compared with the state-of-the-art methods.

\end{abstract}
\vspace{-.05in}

\vspace{-.1in}
\section{Introduction}
\label{sec:intro}
\vspace{-.1in}
Anomaly detection has been a long-standing task across diverse applications including network security, medical diagnostics, and video surveillance. In recent years, unsupervised anomaly detection and localization have been extensively studied and widely applied in defect detection of different kinds of industrial products.

 Existing methods can be divided into two categories. The first one attempts to compute the outlier points as anomalies \cite{Defard2020PaDiMAP}\cite{Wan2022IndustrialIA}\cite{Tien2023RevisitingRD}\cite{Hendrycks2018DeepAD}. MKD \cite{Salehi2020MultiresolutionKD} and RD++ \cite{Tien2023RevisitingRD} derive the anomalous regions by comparing the features extracted by a pretrained backbone with another backbone trained using distillation learning with the normal samples. The second category attempts to reconstruct the images according to the normal patterns and then compare them with the input images for anomaly detection~\cite{Liang2022OmniFrequencyCR}\cite{Zavrtanik2021DRMA}\cite{Zhang2022PrototypicalRN}\cite{Wyatt2022AnoDDPMAD}. The auto encoder is one of the current mainstreams.
 ARNet~\cite{Ye2019AttributeRF} proposes an Attribute Clearing Module to erase the attributes associated with the surface semantic representations and reconstruct the images through the auto encoder.
 Draem~\cite{Zavrtanik2021DRMA} utilizes an auto encoder for image reconstruction and a U-Net\cite{Ronneberger2015UNetCN} to generate an anomaly map. OCR-GAN~\cite{Liang2022OmniFrequencyCR} divides an image into multiple frequency bands, reconstructs them using different auto encoders, and then calculates the anomalous regions using a discriminator. PRN~\cite{Zhang2022PrototypicalRN} proposes multi-scale prototypes and multi-scale self-attention to optimize the skip connections in U-Net~\cite{Ronneberger2015UNetCN}.

\begin{figure}[t]
  \begin{minipage}[b]{1.0\linewidth}
  \centering
    \begin{subfigure}{0.48\textwidth}
        \includegraphics[width=\linewidth]{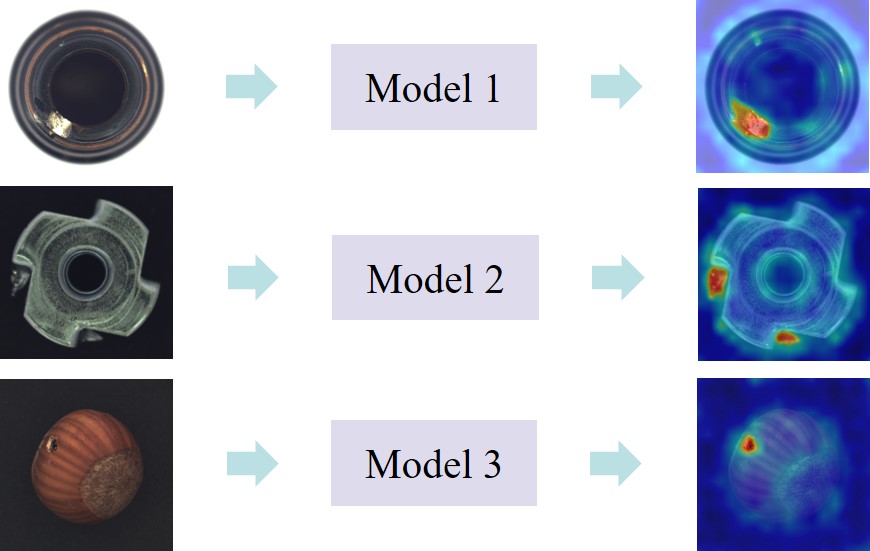}
        \caption{separate paradigm}
    \end{subfigure}
    \begin{subfigure}{0.48\textwidth}
        \includegraphics[width=\linewidth]{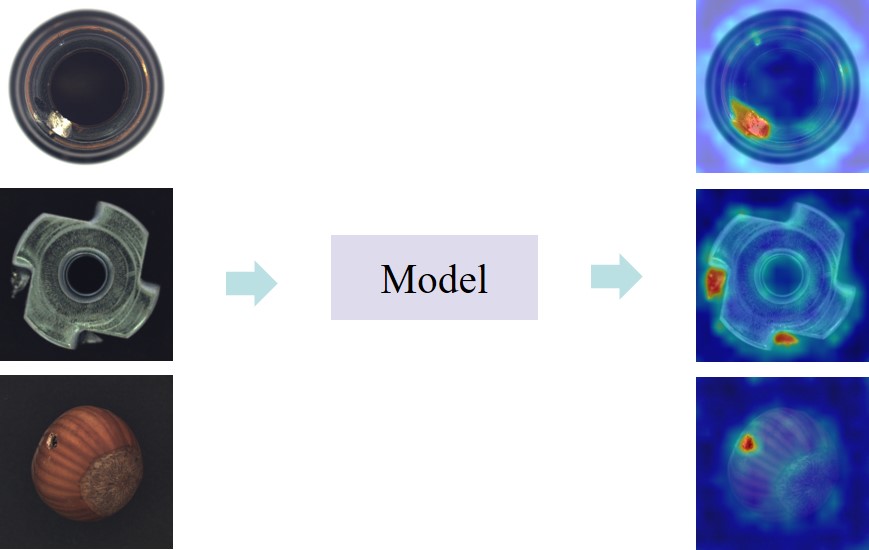}
        \caption{unified paradigm}
    \end{subfigure}
    \vspace{-.15in}
    \caption{Single-class vs. multi-class anomaly detection.}
    \label{fig:fig1}
      \vspace{-.08in}
  \end{minipage}

  \end{figure}

  \begin{figure}[t]
  \begin{minipage}[b]{1.0\linewidth}
  \centering
  \begin{subfigure}{0.22\textwidth}
    \includegraphics[width=\linewidth]{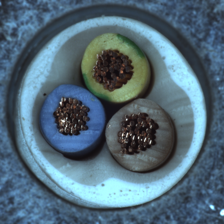}
\end{subfigure}
\begin{subfigure}{0.22\textwidth}
    \includegraphics[width=\linewidth]{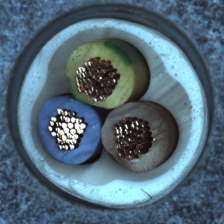}
\end{subfigure}
\begin{subfigure}{0.22\textwidth}
    \includegraphics[width=\linewidth]{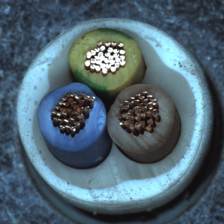}
\end{subfigure}
\begin{subfigure}{0.22\textwidth}
    \includegraphics[width=\linewidth]{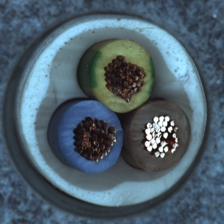}
\end{subfigure}
\vspace{-.1in}
\end{minipage}
\caption{The images of the same normal object under different lighting conditions. Some color channels are distorted and the objects might be falsely detected as anomalies.}
\vspace{-.2in}
\label{fig:fig4}
\end{figure}

  

\begin{figure*}[t]
  \centering
  \begin{minipage}{0.7\textwidth}
    \centering
    \includegraphics[width=\linewidth]{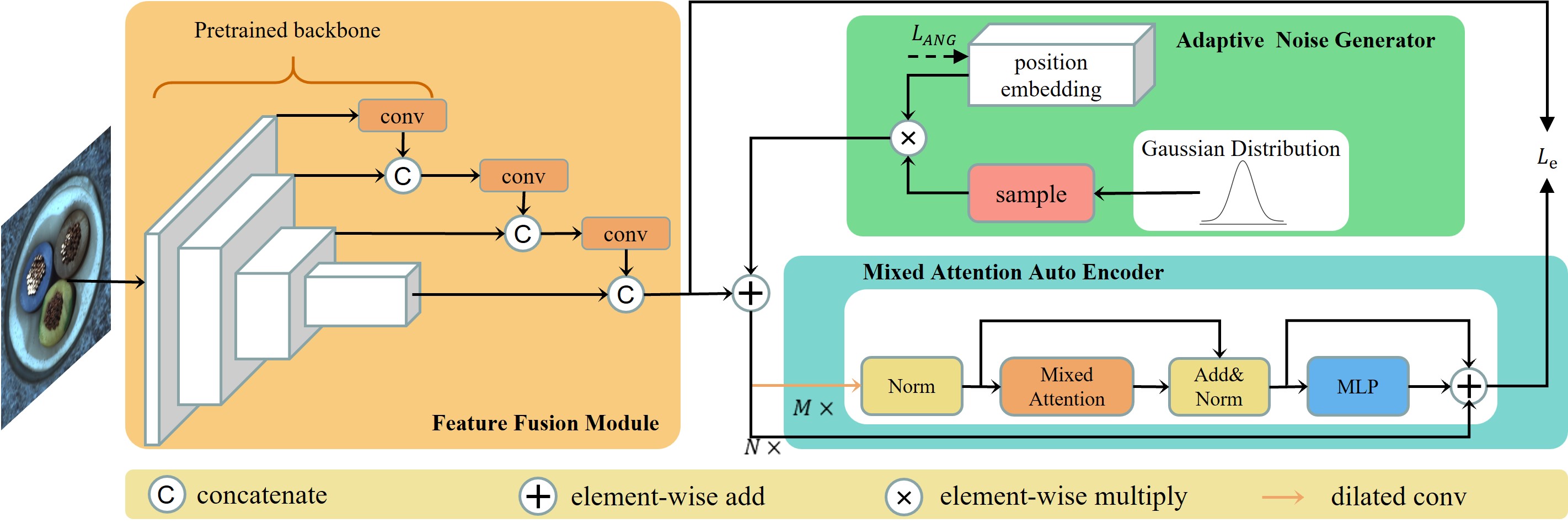}
    \caption*{(a)}
  \end{minipage}%
  \begin{minipage}{0.3\textwidth}
    \centering
    \includegraphics[width=\linewidth]{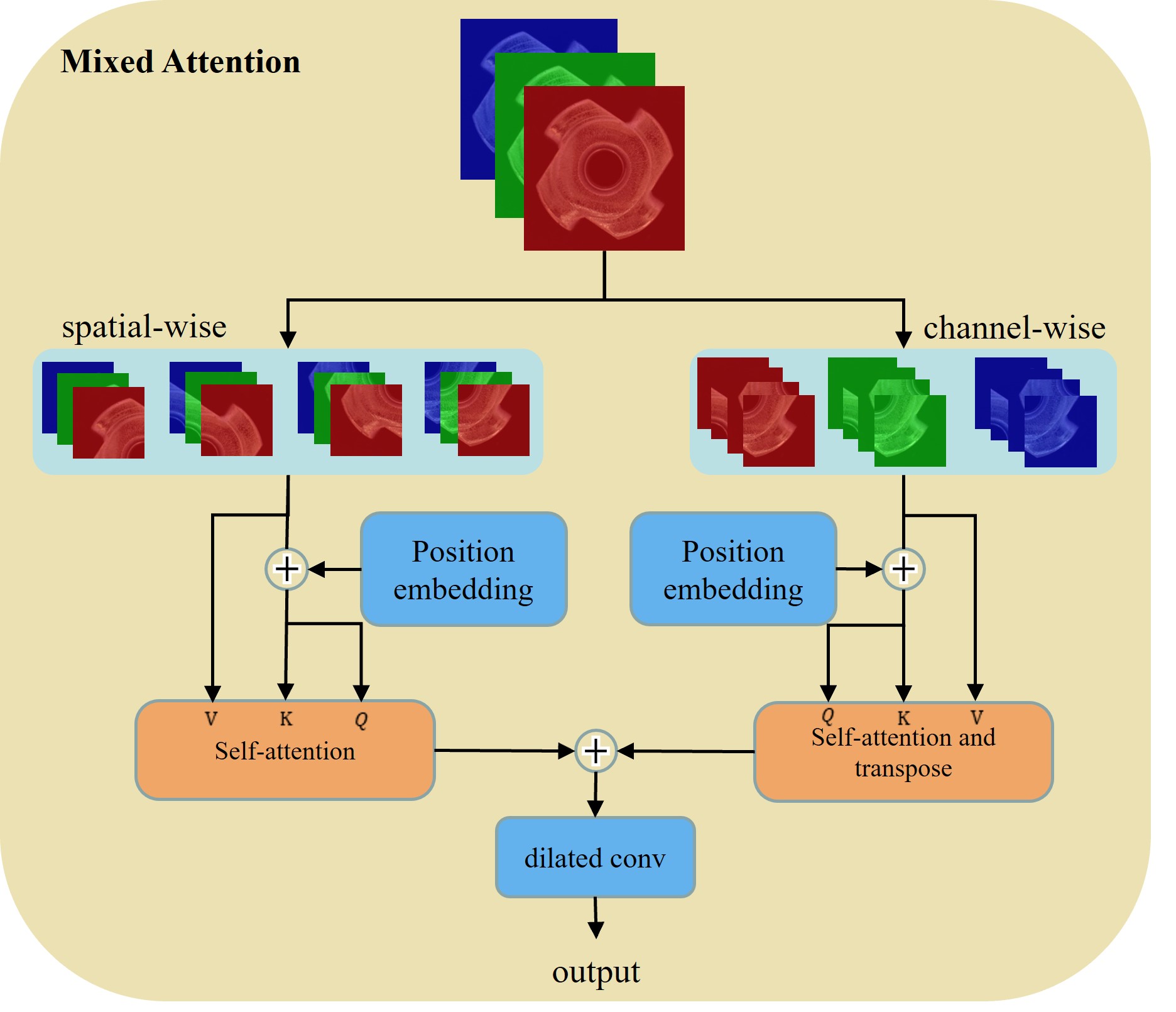}
      \vspace{-.3in}
    \caption*{(b)}
  \end{minipage}
  \vspace{-.15in}
  \caption{(a) The overall architecture of our approach. (b) Illustration of the proposed mixed-attention module.}
  \label{fig:fig2}
    \vspace{-.15in}
\end{figure*}

 These methods have achieved promising performances on the real-world datasets.
However, as shown in Fig.~\ref{fig:fig1}(a), most existing methods need to train a separate model for each class of items. This leads to high storage cost and low training efficiency. An alternative way is to train a unified model which is able to detect anomalies from objects of different classes as shown in Fig.~\ref{fig:fig1}(b). However, this often leads to a serious degradation in performance. Since different objects have different spatial distribution patterns, it is hard for the existing methods to simultaneously model the feature distributions of multiple classes with a unified model. UniAD \cite{You2022AUM} constructs a transformer for multi-class anomaly detection. Mask attention is incorporated to fit the complex distributions of multiple classes, and a layer-wise query embedding decoder is used to restore the anomalies. The mask attention emphasizes the global spatial distributions, but has limitations in handling details in lighting or colors which may cause erroneous judgement of normal regions. Fig.~\ref{fig:fig4} shows an example where the same normal object exhibits different color styles under different lighting conditions. By using only the spatial attention, the images might be falsely detected as anomalies. 

In this paper, we propose a unified mixed-attention auto encoder (MAAE) for multi-class anomaly detection. To effectively model the diverse distributions of multiple classes, we employ both the spatial-wise and the channel-wise attentions, and incorporate information through dilated convolution to refine the image reconstruction by the global shapes and the details of the objects to eliminate the effects of various lighting conditions. Additionally, to alleviate the semantic blur problem and appropriately preserve the surface semantics of the subtle anomalies, we incorporate a feature fusion module (FFM) to continuously adjust the low-level features as the higher-level features are added. Finally, to focus on more sensitive channels and patches of different categories during image reconstruction, we propose an adaptive noise generator (ANG) for the auto encoder. 


\begin{table*}[t]
\renewcommand{\arraystretch}{0.95}
\setlength{\tabcolsep} {2.8mm}
     \centering
     \small{
  \begin{tabular}{c|c|c|c|c|c|c||c|c}
  \hline
    \multicolumn{2}{c|}{\multirow{2}{*}{classes}} & \multicolumn{5}{c||}{single-class model} & \multicolumn{2}{c}{multi-class model}\\
    \cline{3-9}
    \multicolumn{2}{c|}{} & PaDim\cite{Defard2020PaDiMAP} & CutPaste\cite{Li2021CutPasteSL} & MKD\cite{Salehi2020MultiresolutionKD} & Draem\cite{Zavrtanik2021DRMA} & RD++\cite{Tien2023RevisitingRD} &UniAD\cite{You2022AUM} & Ours\\
    \hline
     \multirow{5}*{Texture}& Carpet &93.8/\textcolor{gray}{99.8}&93.6/\textcolor{gray}{93.9}&69.8/\textcolor{gray}{79.3}&98.0/\textcolor{gray}{97.0}&98.2/\textcolor{gray}{\textbf{100.0}}&99.8/\textcolor{gray}{99.9}&\textbf{100.0}/\textcolor{gray}{99.9}\\
     & Grid &73.9/\textcolor{gray}{96.7}&93.2/\textcolor{gray}{100.0}&83.8/\textcolor{gray}{78.0}&\textbf{99.3}/\textcolor{gray}{99.9}&90.5/\textcolor{gray}{\textbf{100.0}}&98.2/\textcolor{gray}{98.5}&98.5/\textcolor{gray}{96.6}\\
     & Leather &99.9/\textcolor{gray}{100.0}&93.4/\textcolor{gray}{100.0}&93.6/\textcolor{gray}{95.1}&98.7/\textcolor{gray}{100.0}&100.0/\textcolor{gray}{100.0}&100.0/\textcolor{gray}{100.0}&\textbf{100.0}/\textcolor{gray}{\textbf{100.0}}\\
     & Tile &93.3/\textcolor{gray}{98.1}&88.6/\textcolor{gray}{94.6}&89.5/\textcolor{gray}{91.6}&99.8/\textcolor{gray}{99.6}&98.5/\textcolor{gray}{\textbf{99.7}}&99.3/\textcolor{gray}{99.0}&\textbf{100.0}/\textcolor{gray}{98.7}\\
     & Wood &98.4/\textcolor{gray}{99.2}&80.4/\textcolor{gray}{99.1}&93.4/\textcolor{gray}{94.3}&\textbf{99.8}/\textcolor{gray}{99.1}&98.9/\textcolor{gray}{\textbf{99.3}}&98.6/\textcolor{gray}{97.9}&96.9/\textcolor{gray}{98.0}\\
    \hline
     \multirow{10}*{Object}& Bottle &97.9/\textcolor{gray}{99.9}&67.9/\textcolor{gray}{98.2}&98.7/\textcolor{gray}{99.4}&97.5/\textcolor{gray}{99.2}&99.7/\textcolor{gray}{100.0}&99.7/\textcolor{gray}{\textbf{100.0}}&\textbf{99.9}/\textcolor{gray}{98.9}\\
     & Cable &70.9/\textcolor{gray}{92.7}&69.2/\textcolor{gray}{81.2}&78.2/\textcolor{gray}{89.2}&57.8/\textcolor{gray}{91.8}&88.6/\textcolor{gray}{\textbf{99.2}}&95.2/\textcolor{gray}{97.6}&\textbf{97.4}/\textcolor{gray}{96.8}\\
     & Capsule &73.4/\textcolor{gray}{91.3}&63.0/\textcolor{gray}{98.2}&68.3/\textcolor{gray}{80.5}&65.3/\textcolor{gray}{98.5}&80.7/\textcolor{gray}{\textbf{99.0}}&86.9/\textcolor{gray}{85.3}&\textbf{93.0}/\textcolor{gray}{89.5}\\
     & Hazelnut &85.5/\textcolor{gray}{92.0}&80.9/\textcolor{gray}{98.3}&97.1/\textcolor{gray}{98.4}&93.7/\textcolor{gray}{100.0}&\textbf{100.0}/\textcolor{gray}{\textbf{100.0}}&99.8/\textcolor{gray}{99.9}&99.6/\textcolor{gray}{99.9}\\
     & Metal Nut &88.0/\textcolor{gray}{98.7}&60.0/\textcolor{gray}{99.9}&64.9/\textcolor{gray}{73.6}&72.8/\textcolor{gray}{98.7}&99.8/\textcolor{gray}{\textbf{100.0}}&99.2/\textcolor{gray}{99.0}&\textbf{99.8}/\textcolor{gray}{99.8}\\
     & Pill &68.8/\textcolor{gray}{93.3}&71.4/\textcolor{gray}{94.9}&79.7/\textcolor{gray}{82.7}&82.2/\textcolor{gray}{\textbf{98.9}}&92.6/\textcolor{gray}{98.4}&93.7/\textcolor{gray}{88.3}&\textbf{95.7}/\textcolor{gray}{92.0}\\
     & Screw &56.9/\textcolor{gray}{85.8}&85.2/\textcolor{gray}{88.7}&75.6/\textcolor{gray}{83.3}&\textbf{92.0}/\textcolor{gray}{93.9}&84.2/\textcolor{gray}{\textbf{98.9}}&87.5/\textcolor{gray}{91.9}&89.1/\textcolor{gray}{88.8}\\
     & Toothbrush&95.3/\textcolor{gray}{96.1}&63.9/\textcolor{gray}{99.4}&75.3/\textcolor{gray}{92.2}&90.6/\textcolor{gray}{100.0}&\textbf{99.4}/\textcolor{gray}{\textbf{100.0}}&94.2/\textcolor{gray}{95.0}&89.7/\textcolor{gray}{93.4}\\
     & Transistor &86.6/\textcolor{gray}{97.4}&57.9/\textcolor{gray}{96.1}&73.4/\textcolor{gray}{85.6}&74.8/\textcolor{gray}{93.1}&92.5/\textcolor{gray}{98.5}&\textbf{99.8}/\textcolor{gray}{\textbf{100.0}}&99.4/\textcolor{gray}{99.5}\\
     & Zipper &79.7/\textcolor{gray}{90.3}&93.5/\textcolor{gray}{99.9}&87.4/\textcolor{gray}{93.2}&\textbf{98.8}/\textcolor{gray}{\textbf{100.0}}&96.7/\textcolor{gray}{98.6}&95.8/\textcolor{gray}{96.7}&98.6/\textcolor{gray}{95.2}\\
    \hline
    \multicolumn{2}{c|}{Average} &84.2/\textcolor{gray}{95.5}&77.5/\textcolor{gray}{96.1}&81.9/\textcolor{gray}{87.8}&88.1/\textcolor{gray}{98.0}&94.7/\textcolor{gray}{\textbf{99.4}}&96.5/\textcolor{gray}{96.6}&\textbf{97.2}/\textcolor{gray}{96.9}\\
    \hline
  \end{tabular}
  }
  \vspace{-.1in}
  \caption{AUROC scores of different models for anomaly detection on MVTec-AD dataset. The results of the separate/\textcolor{gray}{unified} paradigms are presented. For the unified training, the evaluations are performed without any fine-tuning.}
  \label{tab:detection}
  \vspace{-.15in}
\end{table*}

  \vspace{-.1in}
\section{PROPOSED METHOD}
  \vspace{-.1in}
\subsection{Feature Fusion Module}
  \vspace{-.1in}
Fig~\ref{fig:fig2}(a) illustrates the overall structure of our proposed approach. Given an input image, we use a pretrained backbone for feature extraction.
The deep semantic embodies the category information, while the surface semantic captures the details. For most subtle anomalies, the high-level information of an item tends to be normal, which may blur out the low-level features of the anomalies. Thus, the features by directly downsampling with bilinear interpolation and concatenating will tend to be normal and thus miss some small anomalies. To address the semantic blur problem, we employ a feature fusion module (FFM) to effectively incorporate the low-level embeddings into the high-level features. In this way, we pay more attentions to the surface anomalies while utilizing the deep semantic to distinguish the category of the item.  

As illustrated in Fig.~\ref{fig:fig2}(a), we utilize the overlapping strided convolutions for downsampling and concatenate them with the high-level features. The concatenated features are treated as new low-level features, and this process is repeated iteratively. We downsample the low-level features using the convolutional layers with a 3 $\times$ 3 kernel, stride 2, and preserve the original number of channels. Since the neighborhood information has been consistently aggregated in this module, we aggregate the information from distant parts through dilated convolutions before the reconstruction.

 \vspace{-.1in}
\subsection{Adaptive Noise Generator}
 \vspace{-.06in}
As discussed in~\cite{Vincent2008ExtractingAC}, during the training of the auto encoder, if the expected output is always the same as the input, it will encourage the model to learn the identity shortcut (i.e. $ f(x) = x $, where $f$ is our model and $x$ is the input feature). Therefore, an appropriate noise generator is essential for restoring the anomalies. Most previous approaches\cite{Park2023TwostreamDF}\cite{Defard2020PaDiMAP}\cite{Li2021CutPasteSL} add noises by randomly replacing specific regions of the original image. 
These priori noises might be beneficial for reconstructing the items of some specific categories, but may be too strong for other categories and distort the image reconstruction.

To generate noises that effectively work for different categories, we propose an adaptive noise generator (ANG). First, we initialize a weight matrix $W$ of size $N \times C$, where N is the number of patches and $C$ is the dimension of features. Next, we sample noise $\epsilon$ from a Gaussian distribution $N(0,1)$. Finally, the noise $\epsilon'$ is obtained by performing the matrix multiplication $\epsilon' = A \times W \odot \epsilon$, where $\odot$ is element-wise multiplication of matrices, and A is a hyperparameter to control the intensity of the noise. Given an input feature $X$, we add the noise and get the output $X^*=X+\epsilon'$.

Our purpose is to suppress the convergence of the model, and the loss function of this portion is defined as
\begin{equation}
    L_{ANG}=-\lambda_{ANG}L_e+\lambda_{re} \|\mathbf{w}\|_2
    \label{eq:noiseloss}
\end{equation}
where $\lambda_{ANG}$ and $\lambda_{re}$ are hyperparameters, $L_e$ is the loss function of the reconstruction module. If the noise is too large, the model will not be able to accurately reconstruct the anomalies and fail to learn the underlying distributions from the samples. 
Therefore, we introduce L2 regularization in Equation~\ref{eq:noiseloss} to mitigate the expansion of the weight matrix and maintain its stability. 

With the adaptive noise generator, we can generate noises in the areas and the channels which are critical to the normal features. Compared to other noise generators, ANG learns adaptive weights for multi-class objects and could control the intensities of the noises to more flexibly fit the normal samples for accurate image reconstruction.

\vspace{-.1in}
\subsection{Mixed-Attention Auto Encoder}
\vspace{-.08in}
\label{ssec:maae}
From a global perspective, the deep semantics of items within the same category are highly similar. The attention mechanism can help approximate the category information of items. However, as illustrated in Fig.~2, some patches in images may exhibit distortions in certain channels due to lighting or shadows. With the spatial-wise self-attention, these local distortions could be considered as out-of-distribution cases and result in false alarms in anomaly detection. To address this problem, we propose a mixed-attention module as illustrated in Fig.~\ref{fig:fig2}(b) which further employs channel-wise self-attention to recover the local distortions for accurate image reconstruction by making use of the information from the normal channels.

First, we compute the output of the spatial-wise self-attention $Y^{N \times C}_s=SA(X)$ based on the input $X^{N \times C}$. Then, the input is transposed to get the channel distribution $X^{C \times N}$, which is fed to the channel-wise self-attention. We transpose the output of the channel-wise self-attention $Y^{C \times N}_c$ to restore it back to the spatial distribution and add it to $Y_s$. Finally, the combined output is passed through a dilated convolution layer (DC). Since the distant features usually have less influence on reconstructing the local features and the proximal features encapsulate more local information, we aggregate information from intermediate distances to effectively capture the details. This process is presented as
\vspace{-.08in}
\begin{equation}
    Y=DC({SA(X)+(SA(X^T))^T}).
    \label{eq:mixed-attention}
    \vspace{-.08in}
\end{equation}
Furthermore, We add a residual structure after every $M$ blocks to prevent information loss. In terms of the loss function, the difference between the denoised feature $Y$ and the original feature $X^*$ is computed using the Mean Squared Error loss
\begin{equation}
    L_e=\frac{||Y-X^*||_2}{N}.
    \label{eq:loss}
    \vspace{-.05in}
\end{equation}
The anomaly map is upsampled by the difference between the restored vector and the original vector
\vspace{-.05in}
\begin{equation}
    S_l=\ upsample(||Y-X^*||_2),
    \label{eq:segment}
    \vspace{-.05in}
\end{equation} 
and the anomaly score is calculated as the maximum value in the anomaly map
\vspace{-.05in}
\begin{equation}
    S_d=max(S_l).
    \label{eq:score}
    \vspace{-.05in}
\end{equation}

\section{EXPERIMENTS}

\begin{table*}[t]
\renewcommand{\arraystretch}{0.95}
\setlength{\tabcolsep} {3.4mm}
     \centering
\small{
  \begin{tabular}{c|c|c|c|c|c|c||c|c}
   \hline
    \multicolumn{2}{c|}{\multirow{2}{*}{classes}} & \multicolumn{5}{c||}{single-class model} & \multicolumn{2}{c}{multi-class model}\\
    \cline{3-9}
    \multicolumn{2}{c|}{} & PaDim\cite{Defard2020PaDiMAP} & FCDD\cite{Liznerski2020ExplainableDO} & MKD\cite{Salehi2020MultiresolutionKD} & Draem\cite{Zavrtanik2021DRMA}& RD++\cite{Tien2023RevisitingRD} &UniAD\cite{You2022AUM}& Ours\\
    \hline
     \multirow{5}*{Texture}& Carpet & 97.6/\textcolor{gray}{99.0} & 68.6/\textcolor{gray}{96.0} & 95.5/\textcolor{gray}{95.6} & 98.6/\textcolor{gray}{95.5} & \textbf{99.1}/\textcolor{gray}{\textbf{99.2}}& 98.5/\textcolor{gray}{98.0}   & 98.6/\textcolor{gray}{97.8}\\
     & Grid & 71.0/\textcolor{gray}{97.1} & 65.8/\textcolor{gray}{91.0} & 82.3/\textcolor{gray}{91.8} & \textbf{98.7}/\textcolor{gray}{\textbf{99.7}}& 95.3/\textcolor{gray}{98.3} & 96.5/\textcolor{gray}{94.6}  & 97.2/\textcolor{gray}{96.2}\\
     & Leather &84.8/\textcolor{gray}{99.0}&66.3/\textcolor{gray}{98.0}&96.7/\textcolor{gray}{98.1}&97.3/\textcolor{gray}{98.6}& \textbf{99.5}/\textcolor{gray}{\textbf{99.4}}&98.8/\textcolor{gray}{98.3}  & 98.2/\textcolor{gray}{98.8}\\
     & Tile &80.5/\textcolor{gray}{94.1}&59.3/\textcolor{gray}{91.0}&85.3/\textcolor{gray}{82.8}&\textbf{98.0}/\textcolor{gray}{\textbf{99.2}}& 96.6/\textcolor{gray}{96.5}&91.8/\textcolor{gray}{91.8} & 92.2/\textcolor{gray}{92.0}\\
     & Wood &89.1/\textcolor{gray}{94.1}&53.3/\textcolor{gray}{88.0}&80.5/\textcolor{gray}{84.8}&\textbf{96.0}/\textcolor{gray}{\textbf{96.4}}& 95.7/\textcolor{gray}{95.8} &93.2/\textcolor{gray}{93.4}  & 91.4/\textcolor{gray}{92.0}\\
    \hline
     \multirow{10}*{Object}& Bottle &96.1/\textcolor{gray}{98.2}&56.0/\textcolor{gray}{97.0}&91.8/\textcolor{gray}{96.3}&87.6/\textcolor{gray}{\textbf{99.1}}& 97.8/\textcolor{gray}{98.3}&98.1/\textcolor{gray}{98.1}  & \textbf{98.2}/\textcolor{gray}{98.0}\\
     & Cable &82.0/\textcolor{gray}{96.7}&64.1/\textcolor{gray}{90.0}&89.3/\textcolor{gray}{82.4}&71.3/\textcolor{gray}{94.7}& 87.1/\textcolor{gray}{\textbf{98.4}}&97.3/\textcolor{gray}{96.8}  & \textbf{97.4}/\textcolor{gray}{97.2}\\
     & Capsule &96.9/\textcolor{gray}{98.6}&67.6/\textcolor{gray}{93.0}&88.3/\textcolor{gray}{95.9}&50.5/\textcolor{gray}{94.3}& 98.5/\textcolor{gray}{\textbf{98.8}}&98.5/\textcolor{gray}{97.9}  & \textbf{98.7}/\textcolor{gray}{98.1}\\
     & Hazelnut &96.3/\textcolor{gray}{98.1}&79.3/\textcolor{gray}{95.0}&91.2/\textcolor{gray}{94.6}&96.9/\textcolor{gray}{99.7}& 98.0/\textcolor{gray}{\textbf{99.2}}&98.1/\textcolor{gray}{98.8}  & \textbf{98.1}/\textcolor{gray}{98.2}\\
     & Metal Nut &84.8/\textcolor{gray}{97.3}&57.5/\textcolor{gray}{94.0}&64.2/\textcolor{gray}{86.4}&62.2/\textcolor{gray}{\textbf{99.5}}& 95.4/\textcolor{gray}{98.1}&94.8/\textcolor{gray}{95.7} & \textbf{95.8}/\textcolor{gray}{96.9}\\
     & Pill &87.7/\textcolor{gray}{95.7}&65.9/\textcolor{gray}{81.0}&69.7/\textcolor{gray}{89.6}&94.4/\textcolor{gray}{97.6}& 94.9/\textcolor{gray}{\textbf{98.3}}&95.0/\textcolor{gray}{95.1} & \textbf{95.2}/\textcolor{gray}{95.5}\\
     & Screw &94.1/\textcolor{gray}{98.4}&67.2/\textcolor{gray}{86.0}&92.1/\textcolor{gray}{96.0}&95.5/\textcolor{gray}{97.6}& 98.2/\textcolor{gray}{\textbf{99.7}}&98.3/\textcolor{gray}{97.4} & \textbf{98.5}/\textcolor{gray}{98.7}\\
     & Toothbrush&95.6/\textcolor{gray}{98.8}&60.8/\textcolor{gray}{94.0}&88.9/\textcolor{gray}{96.1}&97.7/\textcolor{gray}{98.1}& 98.1/\textcolor{gray}{\textbf{99.1}}&\textbf{98.4}/\textcolor{gray}{97.8} & 98.2/\textcolor{gray}{98.7}\\
     & Transistor &92.3/\textcolor{gray}{97.6}&54.2/\textcolor{gray}{88.0}&71.7/\textcolor{gray}{76.5}&64.5/\textcolor{gray}{90.9}& 87.1/\textcolor{gray}{94.3}&97.9/\textcolor{gray}{98.7} & \textbf{97.9}/\textcolor{gray}{\textbf{98.7}}\\
     & Zipper &94.8/\textcolor{gray}{98.4}&63.0/\textcolor{gray}{92.0}&86.1/\textcolor{gray}{93.9}&\textbf{98.3}/\textcolor{gray}{98.8}& 97.4/\textcolor{gray}{\textbf{98.8}}&96.8/\textcolor{gray}{96.0} & 97.6/\textcolor{gray}{98.2}\\
    \hline
    \multicolumn{2}{c|}{Average} &89.5/\textcolor{gray}{97.4}&63.3/\textcolor{gray}{92.0}&84.9/\textcolor{gray}{90.7}&87.2/\textcolor{gray}{97.3}& 95.8/\textcolor{gray}{\textbf{98.2}}&96.8/\textcolor{gray}{96.6} & \textbf{96.9}/\textcolor{gray}{96.9}\\
    \hline
  \end{tabular}
  }
  \vspace{-.12in}
  \caption{AUROC scores of different models for anomaly localization on MVTec-AD dataset. The results of the unified/\textcolor{gray}{separate} paradigms are presented. For the unified training, the evaluations are performed without any fine-tuning.}
  \label{tab:segmentation}
  \vspace{-.15in}
\end{table*}

\vspace{-.1in}
\subsection{Experimental Setup}
\vspace{-.08in}
\textbf{Implementation Details.} We adopt EfficientNet-b4\cite{tan2019efficientnet} pretrained on ImageNet\cite{Deng2009ImageNetAL} as the backbone, and fuse features from stage-1 to stage-4. The images are resized to 256 $\times$ 256. We use the Adam optimizer with a learning rate of $10^{-4}$. MAAE is trained in both the unified and the separate paradigms with a batch size of $64$. The weights $\lambda_{ANG}$ and $\lambda_{re}$ in Equation~\ref{eq:noiseloss} are empirically set to $0.6$ and $1$ respectively. The dilation rate of the dilated convolution is set to $4$. The number of encoder layers is $18$, and $M$ in section \ref{ssec:maae} is set to $3$. Our experiments are conducted on one NVIDIA RTX 3090 GPU.

\noindent\textbf{Dataset.} MVTec-AD\cite{Bergmann2019MVTecA} is a benchmark dataset widely used for industrial anomaly detection. It contains over 5,000 high-resolution images of 15 different object and texture categories. Each category comprises a set of defect-free training images and a test set of images with various kinds of defects as well as images without defects. 

\noindent\textbf{Evaluation metric.} We use the Area Under the Receiver Operating Characteristic Curve(AUROC) on the image level and the pixel level for performance evaluation which is the most widely adopted in anomaly detection. 
\label{subsec:ablation}
\begin{table}[t]
\renewcommand{\arraystretch}{0.95}
\setlength{\tabcolsep} {3.4mm}
    \centering
    \small{
    \begin{tabular}{ccc|c|c}
         \hline
          \multirow{2}*{ANG} & \multirow{2}*{FFM} & \multirow{2}*{MAAE} & AUROC & AUROC\\
           & & &image-wise &pixel-wise\\
          \hline
           - &- &- &76.0 &88.4\\
           - &- &\checkmark &76.9 &89.4\\
           \checkmark &- &- &92.3 &95.2\\
           \checkmark &\checkmark &- &96.9 &96.6\\
           \checkmark &- &\checkmark &93.9 &96.8\\
           \checkmark &\checkmark &\checkmark &\textbf{97.2} &\textbf{96.9}\\
          \hline
    \end{tabular}
    }
    \vspace{-.1in}
    \caption{Ablation study of different modules in our approach.} 
    \label{tab:ablation}
    \vspace{-.15in}
\end{table}

\vspace{-.1in}
\subsection{Experimental Results}
\vspace{-.1in}
 Tables~\ref{tab:detection} and~\ref{tab:segmentation} show the quantitative results of anomaly detection and localization conducted on different categories of the MVTec-AD dataset. The preceding five models are originally designed for single-class anomaly detection, whereas UniAD\cite{You2022AUM} is a unified model for multi-class anomaly detection. For fair comparison, we evaluate each model twice by following the separate and the unified paradigms respectively. 
 
\begin{figure}[t]
  \begin{minipage}[b]{1.0\linewidth}
  \centering
  \begin{subfigure}{0.15\textwidth}
    \includegraphics[width=\linewidth]{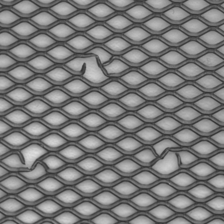}
\end{subfigure}
\begin{subfigure}{0.15\textwidth}
    \includegraphics[width=\linewidth]{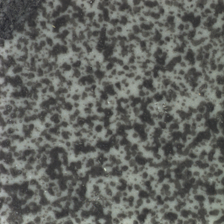}
\end{subfigure}
\begin{subfigure}{0.15\textwidth}
    \includegraphics[width=\linewidth]{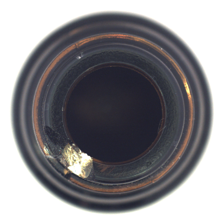}
\end{subfigure}
\begin{subfigure}{0.15\textwidth}
    \includegraphics[width=\linewidth]{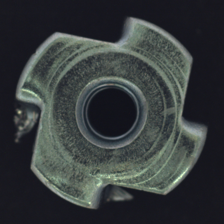}
\end{subfigure}
\begin{subfigure}{0.15\textwidth}
    \includegraphics[width=\linewidth]{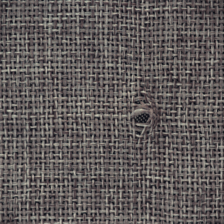}
\end{subfigure}
\begin{subfigure}{0.15\textwidth}
    \includegraphics[width=\linewidth]{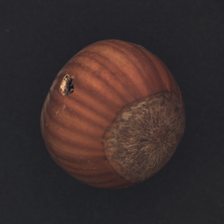}
\end{subfigure}
\begin{subfigure}{0.15\textwidth}
    \includegraphics[width=\linewidth]{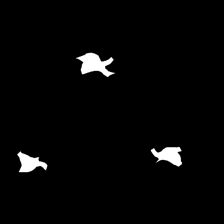}
\end{subfigure}
\begin{subfigure}{0.15\textwidth}
    \includegraphics[width=\linewidth]{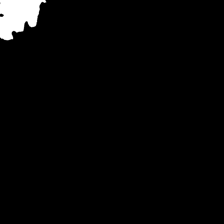}
\end{subfigure}
\begin{subfigure}{0.15\textwidth}
    \includegraphics[width=\linewidth]{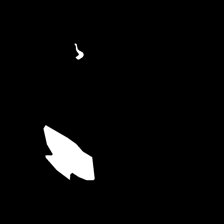}
\end{subfigure}
\begin{subfigure}{0.15\textwidth}
    \includegraphics[width=\linewidth]{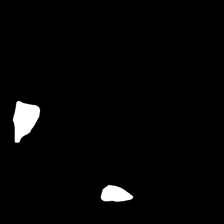}
\end{subfigure}
\begin{subfigure}{0.15\textwidth}
    \includegraphics[width=\linewidth]{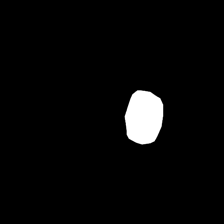}
\end{subfigure}
\begin{subfigure}{0.15\textwidth}
    \includegraphics[width=\linewidth]{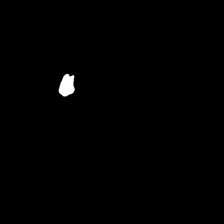}
\end{subfigure}
\begin{subfigure}{0.15\textwidth}
    \includegraphics[width=\linewidth]{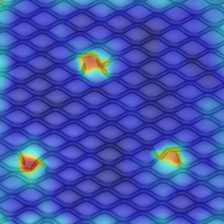}
\end{subfigure}
\begin{subfigure}{0.15\textwidth}
    \includegraphics[width=\linewidth]{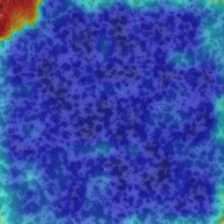}
\end{subfigure}
\begin{subfigure}{0.15\textwidth}
    \includegraphics[width=\linewidth]{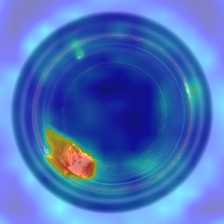}
\end{subfigure}
\begin{subfigure}{0.15\textwidth}
    \includegraphics[width=\linewidth]{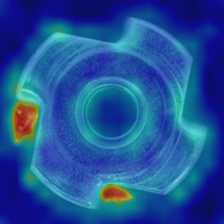}
\end{subfigure}
\begin{subfigure}{0.15\textwidth}
    \includegraphics[width=\linewidth]{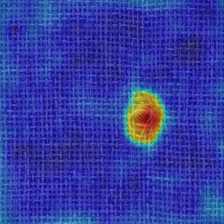}
\end{subfigure}
\begin{subfigure}{0.15\textwidth}
    \includegraphics[width=\linewidth]{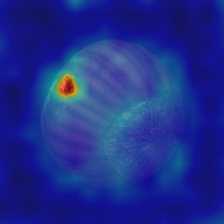}
\end{subfigure}
\end{minipage}
\vspace{-.25in}
\caption{Qualitative results of anomaly localization with our approach. The original images, the groundtruths, and the anomaly heatmaps are shown from top to bottom.}
\label{fig:fig3}
\vspace{-.15in}
\end{figure}

 As can be seen in Tables~\ref{tab:detection} and~\ref{tab:segmentation}, most single-class models suffer significant performance degradations when being used for multi-class anomaly detection. For instance, for categories {\em cable} and {\em transistor}, the performance decline is over $10\%$. In contrast, the average performance of our method does not decrease or is even improved by at most $4\%$ for individual categories. Our approach, after undergoing multi-class training, outperforms the state-of-the-art approaches for more than half of the categories and the average performance. This demonstrates the effectiveness of our approach in modeling the diverse feature distributions of multiple classes. Compared with another multi-class model, our MAAE outperforms UniAD for most object categories and the average performance in both anomaly detection and localization tasks. By combining the spatial and the channel self-attentions, our approach demonstrates more effective in modeling the feature distribution and robust to various lighting conditions. Fig.~\ref{fig:fig3} shows the precise anomaly localization results achieved by MAAE. 

\subsection{Ablation Study}
\vspace{-.1in}
Table~\ref{tab:ablation} illustrates the effects of three modules in our approach. For comparison, we remove ANG, use bilinear downsampling concatenation in place of FFM, and replace our mixed-attention with self-attention \cite{Vaswani2017AttentionIA}. ANG enhances the model's resilience to noises by adaptively simulating the noises of different categories during training. FFM extracts crucial information and refines it to further boost detection performance. Finally, MAAE improves the encoding and decoding capabilities and significantly enhances the localization performance. 


\vspace{-.05in}
\section{Conclusion}
\vspace{-.15in}
We have presented our mixed-attention auto encoder for multi-class industrial anomaly detection. By fusing the low-level and the high-level features, we can eliminate the semantic blur problem in image reconstruction. Our proposed adaptive noise generator is capable of adding noises to the feature-sensitive parts. To address the diverse feature distributions of different categories, we combine the spatial-wise and the channel-wise self-attentions which enable us to detect anomalies from multiple classes with a unified model and alleviate the color distortions caused by various lighting conditions.

\bibliographystyle{IEEEbib}
\newpage
\bibliography{references}

\end{document}